\documentclass[10pt,twocolumn,letterpaper]{article}

\usepackage{cvpr}              

\usepackage{graphicx}
\usepackage{amsmath}
\usepackage{amssymb}
\usepackage{booktabs}
\usepackage{paralist}
\usepackage{multirow}
\usepackage{array}
\usepackage{enumitem}

\usepackage[pagebackref,breaklinks,colorlinks]{hyperref}

\newcommand{\eric}[1]{{\color{red}[Eric]}}

\usepackage[capitalize]{cleveref}
\crefname{section}{Sec.}{Secs.}
\Crefname{section}{Section}{Sections}
\Crefname{table}{Table}{Tables}
\crefname{table}{Tab.}{Tabs.}

\def\cvprPaperID{889} 
\def\confName{CVPR}
\def\confYear{2022}

\begin{document}


\title{Fixing Malfunctional Objects With \\ Learned Physical Simulation and Functional Prediction}

\author{Yining Hong\\
UCLA\\
\and
Kaichun Mo\\
Stanford University\\
\and
Li Yi\\
Tsinghua University\\
\and
Leonidas J. Guibas\\
Stanford University\\
\and
Antonio Torralba\\
MIT\\
\and
Joshua B. Tenenbaum\\
MIT BCS, CBMM, CSAIL\\
\and
Chuang Gan\\
MIT-IBM Watson AI Lab\\
}

\twocolumn[{%
\renewcommand\twocolumn[1][]{#1}%
\maketitle
    \centering
    \includegraphics[width=0.9\textwidth]{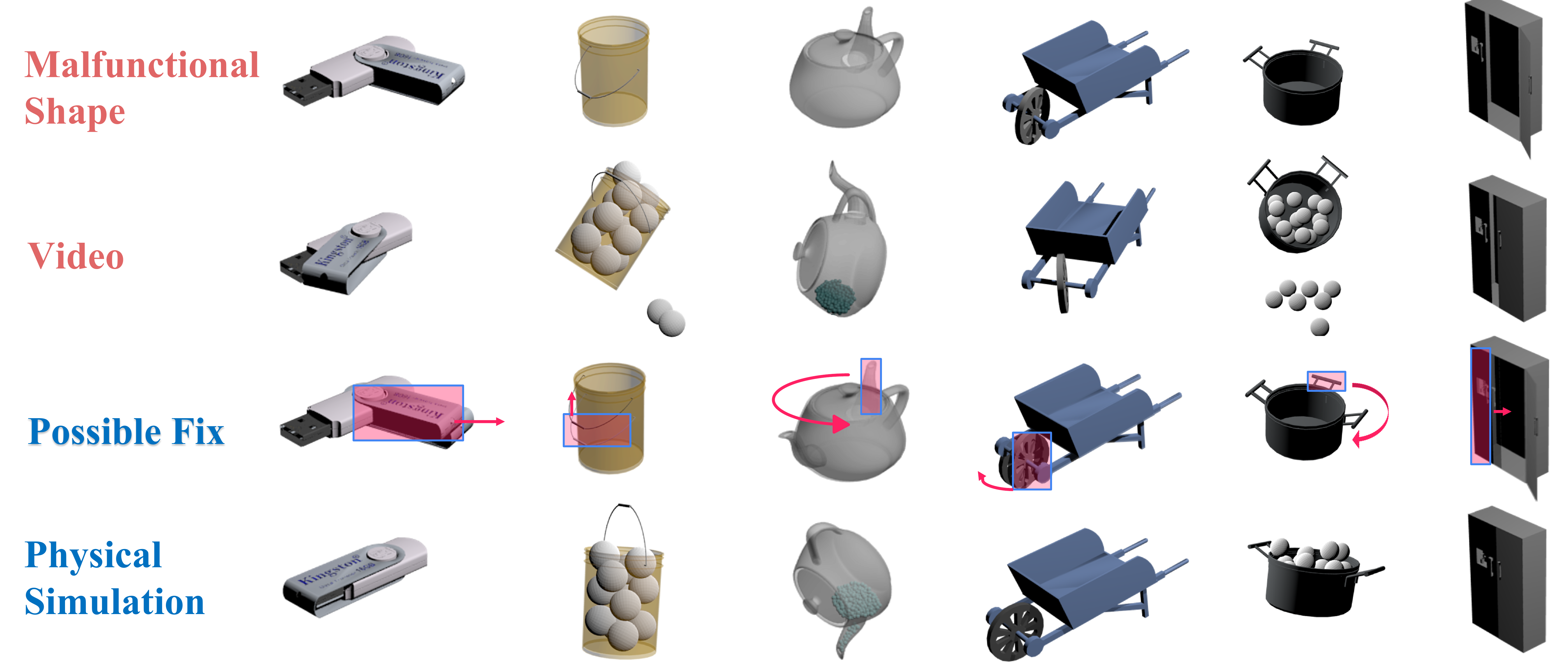}
    \captionof{figure}{We introduce \textsc{FixIt}, a dataset that requires machines to fix malfunctional objects based on functionality. Each malfunctional object is paired with a video presenting how the object is interacted. Functionality of fixed objects can be evaluated via physical simulation. 
    }
    \label{fig:teaser}

\vspace{5mm}
}]

\begin{abstract}
This paper studies the problem of fixing malfunctional 3D objects. While previous works focus on building passive perception models to learn the functionality from static 3D objects, we argue that functionality is reckoned with respect to the physical interactions between the object and the user. Given a malfunctional object, humans can perform mental simulations to reason about its functionality and figure out how to fix it. Inspired by this, we propose \textsc{FixIt}, a dataset that contains about 5k poorly-designed 3D physical objects paired with choices to fix them. To mimic humans' mental simulation process, we present FixNet, a novel framework that seamlessly incorporates perception and physical dynamics. Specifically, FixNet consists of a perception module to extract the structured representation from the 3D point cloud, a physical dynamics prediction module to simulate the results of interactions on 3D objects, and a functionality prediction module to evaluate the functionality and choose the correct fix. Experimental results show that our framework outperforms baseline models by a large margin, and can generalize well to objects with similar interaction types. Code and dataset are publicly available\footnote{\url{ http://fixing-malfunctional.csail.mit.edu}}.
\end{abstract}

\section{Introduction}

What defines a good 3D object shape? Aspects like aesthetics, comfort and fun \textit{etc} play critical roles in shape design.
However, these perspectives are all rendered meaningless without the utmost consideration of an object for everyday use - its functionality. 
Functionality is a relation between the goal of users' interaction and the behavior of the object. Each part of the object relates its behavior to the functionality of the entire object shape. For example, the third column in Figure~\ref{fig:teaser} shows the notorious ``Coffeepot  for Masochists"  by the French artist  Jacques  
Carelman \cite{carelman1969catalogue}, in which the rotation of the spout affects the functionality of the coffeepot.
There have been recent efforts on learning the functionality of 3D object shapes\cite{hu2020predictive, hu2016learning, hu2015interaction, mo2021where2act, mo2021o2o}. These works usually treat the functionality as a property of the 3D objects, and use \textit{perception systems} to predict the property.

This practice is consistent with the early ``ecological theory of perception” which claims that people could simply pick up clues from the world through direct perception and make predictions about functionality \cite{gibson1977theory}. 
However, a more widely-accepted notion is that to evaluate functionality, we need to involve physics which are sometimes unperceivable  \cite{norman2013design, hartson2003cognitive}.
In fact, functionality is reckoned with respect to the \textit{physical interactions} between the object and the user, instead of being a property associated with the shape only. For example, in Figure \ref{fig:teaser}, a person has to interact with the USB to find that the shell fails to protect the chip. 

Fixing designs based on functionality is termed as ``functional reasoning in design" \cite{umeda1997functional}. Specifically, to fix an object, one has to modify a part and interact with the object to verify whether the fixed object exhibits the desired function \cite{stamatis2003failure}. However, in  real-life scenarios, it is unrealistic to try out all possible fixes and interact with all the fixed object shapes. Therefore, humans are inclined to perform mental simulation to simulate the effects of interaction \cite{christensen2009role}. For example, once we have seen how the USB can be interacted, we come up with some fixing ideas and mentally simulate the interaction of the fixed USB. Similar mechanisms have been achieved for machines via dynamics models \cite{chang2016compositional, battaglia2016interaction, li2019propagation, li2018learning}. 
These models are able to simulate future states for a given object and interaction.
Equipping machines with the ability to fix objects based on physical dynamics and functionality benefits many real-world applications. For example, it helps predict the outcome of interactions on the objects and recommend fixes to malfunctional objects. When 3D models are designed for virtual reality (VR), it helps guarantee that these 3D models function well.

Inspired by the above ideas, 
we propose a novel task that requires machines to fix malfunctional objects. To study this problem, we have created a new 3D synthetic dataset, \textsc{FixIt}, which contains approximately 5k synthetic point cloud videos of 3D objects. The point cloud videos present the simulations of how the 3D objects are interacted. Most of the interactions are not successful, indicating that the objects do not function well, while a small set of videos demonstrate successful interactions. We pair each video with five choices indicating how to fix the 3D objects. Only one of the five choices is correct.

 As a first attempt at this challenging task, we propose FixNet, a framework that could learn physical dynamics and functional prediction from 3D point cloud videos. Since previous physical dynamics models require full access to particle states and groupings\cite{li2019learning}, a major challenge to apply them for this new task is how to predict physical dynamics from raw point cloud videos. Our idea is that the point cloud in 3D objects, after going through perception system that provides structured representations and point correspondences, can be compatible with the particles (\textit{i.e.}, small localized objects to which can be ascribed physical properties) in the physical dynamics system. Specifically, FixNet is composed of three modules: the perception module, the physical dynamics prediction module and the functionality prediction module. The perception module has two parts: a) a flow prediction network that takes a point cloud video as inputs, and proposes the flows of the points, which are used as pseudo-labels for training the dynamics module; and b) a segmentation network that takes the point cloud and the predicted flows, and proposes the parts of the objects, which are used for fixing. The point clouds of the parts are modified according to the fixing choices. The dynamics prediction module then takes the fixed point clouds, performs physical simulation and outputs an interacting video. Finally, the last step of the simulated video is fed into the functionality prediction module to evaluate whether the fixed object functions well.

Experiments on the \textsc{FixIt} dataset suggest that our FixNet outperforms several baseline models by a large margin. Moreover, it can generalize well to novel categories with the same interaction type. Model diagnosis and qualitative examples show that the challenge of \textsc{FixIt} lies in providing the accurate segmentations for dynamics prediction. Our contributions can be summarized as follows:
\setdefaultleftmargin{1em}{1em}{}{}{}{}
\begin{compactitem}
\item We propose a novel task of fixing 3D object shapes based on functionality that involves 3D perception, physical and functionality reasoning.
\item We propose a new dataset, \textsc{FixIt}, which contains around 5k object shapes across seven categories for fixing.
\item We propose a modular framework, FixNet, which incorporates perception, physical dynamics and functional prediction for fixing.
\item Experimental results show that our FixNet outperforms baseline models by a large margin.
\end{compactitem}

\section{Related Work}

\noindent\textbf{Functionality Modeling.}
The rich space of 3D object shapes, especially man-made ones in our daily lives, results from the diverse functionalities they need to provide for accomplishing various downstream tasks.
It is therefore an important yet challenging research topic to study shape functionality~\cite{hu2018functionality} and affordance~\cite{gibson2014ecological,hassanin2018visual} as a highly relevant concept.
Previous works have explored learning shape functionality and affordance from human annotations~\cite{do2018affordancenet,deng20213d,tang2021interactive}, 
by watching videos or human demonstrations~\cite{koppula2014physically,zhu2015understanding,fang2018demo2vec,nagarajan2019grounded,nagarajan2020ego}, 
and learning from interaction 
by humans~\cite{grabner2011makes,hu2015interaction,kim2014shape2pose,hu2016learning,pirk2017understanding,hu2020predictive} 
or robot agents~\cite{interaction-exploration,fang2020learning,qin2020keto,mo2021where2act,mo2021o2o}.
Many works~\cite{hu2017learning,wang2019shape2motion,guan_tvcg20,mezghanni2021physically} have also demonstrated the importance of parts and structures for well-functional shapes. However, these works mostly focus on perceiving, modeling, and generating shapes with functionalities.
Our work instead proposes a new problem formulation of diagnosing and fixing malfunctional objects.

\noindent\textbf{Physical Scene Understanding.} 
Physical Reasoning is an important aspect of cognitive reasoning \cite{chen2020cvpr, hong2021lbf, hong2021smart, hong2021vlgrammar}. Recently, researchers have focused on using neural networks to predict physical dynamics \cite{chen1990non, wan2001model, battaglia2016interaction, chang2016compositional, mrowca2018flexible, li2019propagation, li2019learning, zfchen2021iclr, ding2021dynamic, hong2021ptr, li2022contact, lin2022diffskill}. Particle-based dynamic systems have been applied to simulate objects of various materials  \cite{mrowca2018flexible, li2019learning, ummenhofer2019lagrangian, janner2018reasoning, NEURIPS2020_ba484941, ma2021diffaqua}. However, these works usually assume that they have access to all states, clusters and physical properties of the physical systems, which  presents a gap between perception and physics. More often, we are presented with raw, irregularly sampled and variant point clouds. There are also dynamic models over latent representations \cite{finn2017deep, babaeizadeh2017stochastic, hafner2019learning, ha2018recurrent, wu2017learning}. However, these implicit models fail to capture the complex physical properties and thus do not show good performances in predicting future states. \cite{li2020visual} proposes to learn visual priors from images. In contrast, we propose to learn perception from 3D point clouds, and use physical dynamics to fix malfunctional 3D object shapes.

\section{The \textsc{FixIt} dataset}
\begin{table}[ht]
    \centering
    \small
    \vspace{-2mm}
    \begin{tabular}{m{0.09\linewidth}|m{0.06\linewidth}m{0.06\linewidth}c}
        \hline
        Cat. & Shapes & Func.
        &Success Definition    \\
        \hline
        Fridge & 1290 & Close & No Collision; No interior exposed   \\
        Bucket & 624 & Lift & Height Change; No Water Out  \\
        USB & 1096 & Sheild & Chip is not exposed  \\
        Kettle & 213 & Pour & Water can be poured out   \\
        Cart & 654 & Move & Move forward without rotation  \\ Pot & 751 & Lift & Height Change; No Water Out  \\
        Box & 327 & Close & No interior exposed  \\
        \hline
    \end{tabular}
        \vspace{-1mm}
    \caption{Statistics and characteristics of the object categories covered by the \textsc{FixIt} dataset.}
    \label{tab:dataset}
    \vspace{-4mm}
\end{table}
We create a new dataset which contains 4,955 3D object instances
represented as point clouds, called \textsc{FixIt}.
Each object is composed of various parts that can be modified. The objects are paired with point cloud videos showing how the objects are interacted and the dynamic outcomes. Choices indicating possible fixes to the parts of the objects are represented as Domain-Specific Language (DSL).
\subsection{Dataset Design}

\noindent\textbf{Object Categories.}
Our dataset contains 7 object categories: \texttt{Refrigerator}, \texttt{Bucket}, \texttt{USB}, \texttt{Kettle}, \texttt{Cart}, \texttt{KitchenPot}, \texttt{Box}. The 3D models are from the PartNet-Mobility \cite{xiang2020sapien} dataset. We choose these categories since they either have rich articulated parts to interact with (\textit{e.g.}, \texttt{Refrigerator}, \texttt{Box}, \texttt{USB}) or the physical interactions are complex (\textit{e.g.}, \texttt{Bucket}, \texttt{Kettle}, \texttt{kitchenPot}). We purposely break some of the objects by scaling, translating or rotating the parts to make them malfunctional. In Table \ref{tab:dataset}, we define the functionality of each object category. In Figure \ref{fig:teaser}, we show some exemplar objects in our dataset. For more examples and details about each category, please refer to the supplementary material.


\noindent\textbf{Point Cloud Video Generation.}
We use PyBullet \footnote{https://pybullet.org/}  to simulate the  physical interactions for our video dataset, as well as to verify the functionality of the objects. For each object, we use an end effector to interact with the objects (for kitchenpot, we use two because there are two handles). We hard code a pre-defined trajectory of the end effector. We use small balls to replace the water in the buckets, kettles and kitchenpots. After the simulation is done, we check the position and rotation change of each object to evaluate whether the object is functional. We finally extract 10 frames out of all the simulation steps to construct the videos. We use furthest-point-sampling to sample a  point cloud of size 2048 of each frame. We extract 16 interacting points between the end effector and the object, serving as an extra input to tell machines how we can interact with the object.

\begin{figure}
    \centering
    \includegraphics[width=0.9\linewidth]{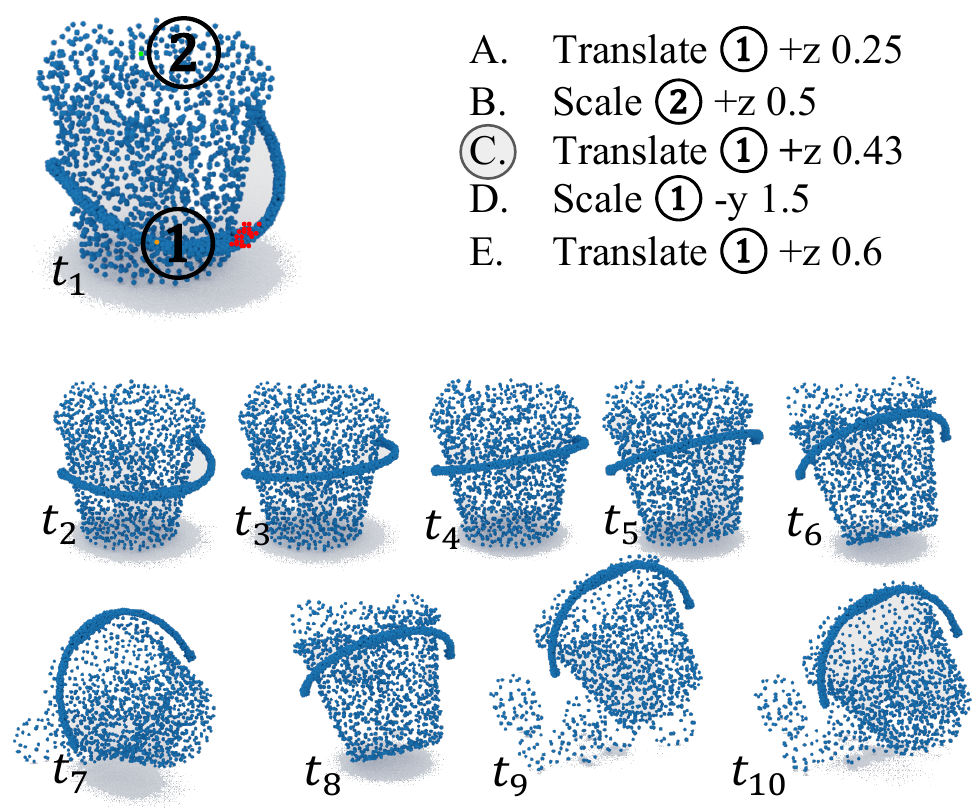}
    \vspace{-2mm}
    \caption{An example of our \textsc{FixIt} dataset. It has several components: 1) 3D point cloud of the shape to be fixed; 2) a point cloud video showing the interaction of this object; 3) root points representing the parts and part indexes; 4) interacting points (the red points); 5) a set of five choices to fix it. Each choice refers to one of the parts via the part index.}
    
    \label{fig:dataset_exp}
    \vspace{-7mm}
\end{figure}

\noindent\textbf{Domain-Specific Language.}
Each fixing choice is represented as a 4-tuple domain-specific language (Type, Part, Axis, Value). There are three fixing types: ``scale", ``translate" and ``rotate"; and six axes: ``+x", ``-x", ``+y", ``-y", ``+z", ``-z". The value is uniformly sampled within a range given the object shape. For each part of the object, we use one root point to represent this part and give it an index. A choice refers to the part to be fixed by specifying the index. A choice can also be  ``functional" indicating the object is already functional and does not need a fix. Figure \ref{fig:dataset_exp} shows an example of choices in dataset.
\begin{figure*}
    \centering
    
    \includegraphics[width=0.9\textwidth]{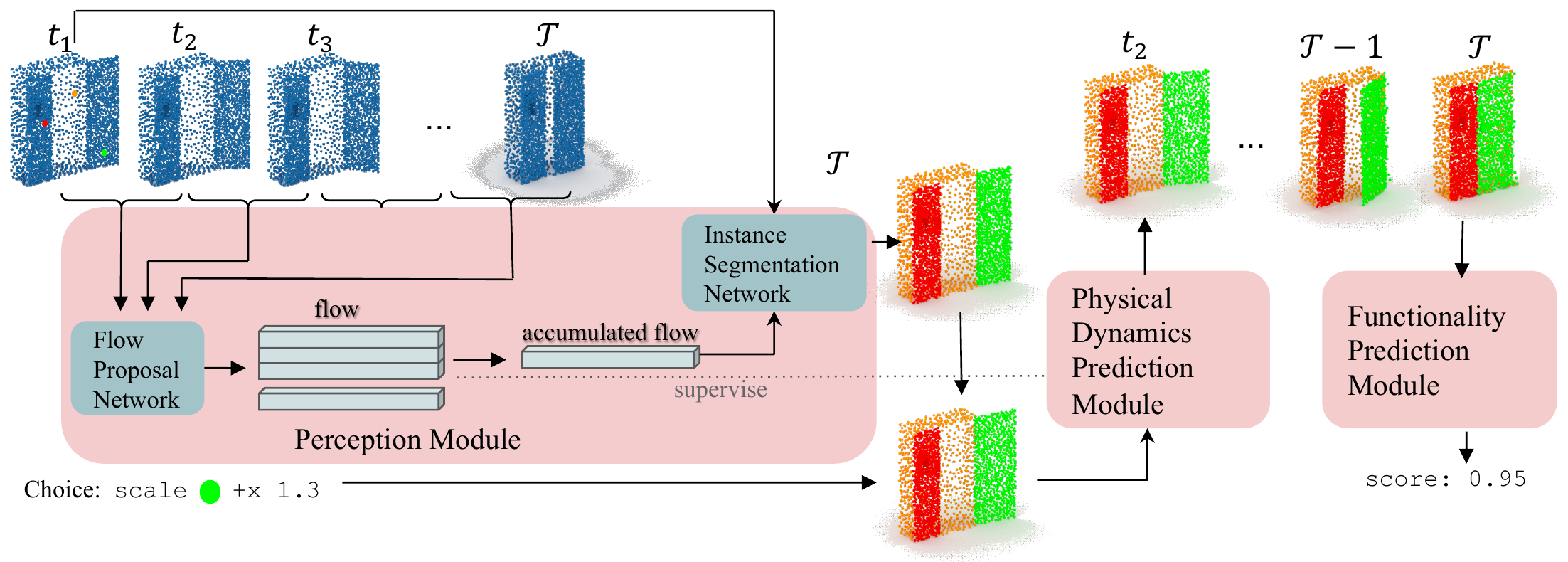}
    \vspace{-3mm}
    \caption{Our proposed FixNet. 
    The point cloud videos are first fed into the flow proposal network which outputs the flows of the points. The flows, and the point cloud of the malfunctional object (which is also the first frame of the video) are input to the segmentation network to produce part instances. Given a choice represented as domain-specific language (DSL), the part referred to in the choice is retrieved via the root point, and modified according to the DSL. The point cloud of the fixed object is then fed into the physical dynamics prediction module, together with the part instances, to predict the future states of the object. The physical dynamics module is trained on the pseudo labels provided by the flows. The last state is input to a functionality prediction module which outputs a functionality score.
    }
    \label{fig:framework}
    \vspace{-4mm}
\end{figure*}
\subsection{Problem Formulation}
For an original point cloud $\mathcal{P}_1$ of the 3D object to be fixed, our framework takes as input a simulated point cloud video $\mathbf{P}=\{\mathcal{P}_{1}, \mathcal{P}_{2},..., \mathcal{P}_{T}\}$, where $T$ denotes the number of frames in the video, which is 10 for our dataset. The point cloud in the $t$-th frame $\mathcal{P}_{i}$ can be represented as a set of points $\mathcal{P}_{i} =  \{p_{t1}, p_{t2},...,p_{tN}\}$ where $N$ equals 2048. We also take in an additional set of interacting points $
\mathcal{IP}_{t} = \{ip_{t1}, ip_{t2},...,ip_{tK}\}$ (K = 16) indicating where the end effector operates on the initial point cloud, and $\mathbf{IP}=\{\mathcal{IP}_1, ..., \mathcal{IP}_T\}$. For an original point cloud $\mathcal{P}_1$ of $L$ parts, we define a set of root point indices $\mathcal{R} = \{r_1, r_2, ..., r_L\}$, where each root is the index in $\mathcal{P}_1$ representing the indicator to a part of the object. Our framework also takes as inputs a set of five choices $\mathcal{C}=\{c_1, c_2,...,c_5\}$ to fix the original object $\mathcal{P}_1$. Each choice refers to the part to be fixed using one of the root points.
\vspace{-2mm}

\section{FixNet}
\vspace{-1mm}
\label{sec:method}

Inspired by humans' mental simulation process, we aim to design AI models that can perform physical simulation of interactions on objects and evaluate the functionality of the fixed objects. There have been works that could predict the dynamics of physical objects accurately \cite{chang2016compositional, battaglia2016interaction, li2019propagation, li2018learning}. However, they require full access to the particle representations, point correspondences and groupings, which are often unobtainable in real-world scenarios. The major challenge resides in learning particle-based dynamics models from raw point cloud videos. Our idea to tackle this challenge is to use the perception module to provide structured representations and point correspondences for the physical dynamics prediction module. Therefore, we present FixNet, a framework that seamlessly bridges the gap between 3D point cloud videos and physical dynamics. As shown in Figure \ref{fig:framework}, our proposed FixNet consists of three modules: a perception module, a physical dynamics prediction module and a functionality prediction module. 
The perception module consists of two networks: a flow proposal network to extract the flows from the point cloud video of the object, and an instance segmentation network to estimate the parts of the object based on the flow. The physical dynamics prediction module takes a segmented object as an input and learns to approximate the physical simulations of its interactions. Finally, the functionality prediction module takes the outcome of simulation and measures if a modified object is well functional or not.

\subsection{3D Visual Perception}
The perception module aims to provide perceptual cues crucial for training the physical dynamics prediction module. It contains two networks: the flow proposal network and the instance segmentation network. 

\noindent\textbf{Flow Proposal Network.}
Since the points in each frame are irregularly sampled, we are unaware of the point correspondences between two frames, thus restricting us from both training the physics simulation module and improving the segmentation network. Therefore, we propose to learn the flow of the points from scratch.

We leverage the scene flow estimation methods \cite{dewan2016rigid, ushani2017learning, behl2019pointflownet, shao2018motion}
to recover flow. Specifically, our flow proposal network is based on FlowNet3D\cite{liu2019flownet3d}, which consists of a PointNet++ to learn embedding, an embedding layer for point mixture and set upconv layers to predict the scene flows.

We re-organize the input point cloud video $\mathbf{P}=\{\mathcal{P}_{1}, \mathcal{P}_{2},..., \mathcal{P}_{T}\}$ into pairs of source points $\mathcal{P}_{t}$ and target points $\mathcal{P}_{t+1}$, where $t=1,2,...,T-1$. The flow proposal network $f_T$ outputs the estimated flow for the source point cloud :
$\Delta \mathcal{ \tilde{P}}_t = \{(\Delta x_{ti}, \Delta y_{ti}, \Delta z_{ti})\}_{i=1}^{N} = f_T(\mathcal{P}_t, \mathcal{P}_{t+1})$.

Note that the estimated flow here is not the actual flow, since $(\mathcal{P}_t + \Delta \mathcal{ \tilde{P}}_t) \neq \mathcal{P}_{t+1}$. To rectify the flow, we compute a $N \times N \times 3$ disparity matrix $\Delta \mathcal{D}_i = \{d_{t}^{jk}\}_{j,k=1}^{N}$ for each pair of points between the $t$-th frame and $t+1$-th frame, where $d_{t}^{jk} = p_{(t+1)k} - p_{tj}$. We expand $\Delta \mathcal{ \tilde{P}}_t$ ($N \times 1 \times 3$) to $[\Delta \mathcal{ \tilde{P}}_t]$ ($N \times N \times 3$), which have the same dim as $\Delta \mathcal{D}_t$, and compute the cost matrix $\mathcal{C}_t=||[\Delta \mathcal{ \tilde{P}}_t]-\mathcal{D}_t||^2$. We apply hungarian algorithm 
\cite{kuhn1955hungarian} on the cost matrix $\mathcal{C}$ to find a bipartite matching $\mathcal{M}_p:\{i \rightarrow \mathcal{M}_p(i) \mid i=1,2, \cdots, N\}$ between the source points $\mathcal{P}_t$ and target points $\mathcal{P}_{t+1}$. This is to minimize the overall error between the estimated flow and the actual flow. We then calculate the rectified flow $\Delta \mathcal{\hat{P}}_t$ from point correspondences:$\Delta \mathcal{\hat{P}}_t = \{p_{(t+1)M(i)} - p_{ti}\}_{i=1}^N$.

\noindent\textbf{Instance Segmentation Network.}
The instance segmentation network proposes part instance segmentations for the succeeding modules. As suggested by \cite{yi2018deep, shi2021self}, the co-segmentation methods which leverage motion flows of articulated objects  achieve good performances. Thus, we devise an instance segmentation network that takes perception information and motion information as inputs and outputs the part instances. We first sum up the flows of all frames as the accumulated flow: $\Delta \mathcal{\hat{P}}=\sum_{t=1}^{T-1}\Delta \mathcal{\hat{P}}_t$.  We concatenate the point cloud of the original object $\mathcal{P}_1$, the accumulated flow to construct the $N \times 6$ inputs for our instance segmentation network $f_{I}$. Our instance segmentation network applies a PointNet++ with multi-scale grouping \cite{qi2017pointnet++} as backbone network for extracting features and predicting $L$ instance segmentation masks over the input point cloud of size $L$: 
$\mathcal{\hat{S}} = \{\hat{s_l} \in [0,1]^{N}|l=1,2,...,L\} = f_{I}([\mathcal{P}_1, \Delta \mathcal{\hat{P}}])$.
A softmax activation layer is applied such that $\hat{s}_{1}+\hat{s}_{2}+...+\hat{s}_{L}=1$. We use Hungarian algorithm to find a bipartite matching $\mathcal{M}_s:\{l \rightarrow \mathcal{M}_s(l) \mid l=1,2, \cdots, L\}$ between the predicted masks $\left\{\hat{s}_{l} \mid l=1,2, \cdots, L\right\}$ and the ground-truth masks $\left\{s_{l} \mid l=1,2, \cdots, L\right\}$. For the metric of Hungarian algorithm, we use a relaxed IOU \cite{krahenbuhl2013parameter}.

\subsection{Physical and Functionality Prediction}

\noindent\textbf{Physical Dynamics Prediction Module.}
We introduce how we utilize the perception prior to simulate physics below. The perception module provides point flows and segmentations that are compatible with particle-based dynamics model \cite{li2018learning, sanchez2020learning, sanchez2018graph}, and the point clouds of the parts referred in the choices to be fixed.  There have been numerous dynamics prediction models proposed recently \cite{chang2016compositional, battaglia2016interaction, li2019propagation}. We chose DPI-Net \cite{li2018learning}, since the particle-based physical dynamics system can naturally leverage the points from the perception systems, and the hierarchical modeling paradigm suits the 3D objects of multiple parts.

The interactions within the physics model can be represented as a directed graph, $G=(\langle \mathcal{P}, E\rangle)$, where $\mathcal{P}$ is the set of points, which are called particles in the physics world, and $E$ is a set of relations between the points. The edges $E$ between particles are dynamically generated over time.

Three types of edges are defined in DPI-Net. The first is to establish relationships among neighbors within a predefined distance. The second type is called hierarchical modeling, where the particles are clustered into non-overlapping clusters, and a random particle in the cluster is selected as root, and other particles are leaf nodes. Directed edges include $E_{LeafToRoot}$, $E_{RootToLeaf}$ and $E_{RootToLeaf}$. DPI-Net employs a multi-stage propagation paradigm: propagation among leaf nodes $\phi_{LeafToLeaf}$; from leaf nodes to roots $\phi_{LeafToRoot}$; between roots $\phi_{RootToRoot}$ and root to leaf $\phi_{RootToLeaf}$. We use the segmentation results $\hat{\mathcal{S}}$ as our clusters and use our root points $\mathcal{R}$ as the root particles. The third type of edges is designed for control. We follow the implementation of \cite{li2018learning}, in which the control inputs are also vertices of the interaction graph, and have directed edges to the points controlled (in real implementation, they choose the points that are close to the positions of the controlling objects as points to be controlled). The dynamics of the interacting points are pre-defined, and only the dynamics of the particles $\mathcal{P}$ are predicted. 

We input the point cloud $\mathcal{P}_{1}$, together with the interacting points $\mathcal{IP}$ for control and the segmentations $\hat{\mathcal{S}}$ for hierarchical modeling. DPI-Net $f_P$ predicts future trajectory of the physical interaction, $\mathbf{{\hat{P}}}=\{\mathcal{{\hat{P}}}_{2},..., \mathcal{{\hat{P}}}_{T}\}$ step by step given an initial object: $\mathbf{{\hat{P}}} = f_P(\mathcal{P}_{1}, \mathcal{\hat{S}}, \mathbf{IP})$.

\noindent\textbf{Functionality Prediction Module.}
The functionality prediction module finally takes in the last frame output by the physical dynamics prediction module and predicts its functionality score by examining whether the goal  of the interaction has been achieved.
\subsection{Training and Inference}
\begin{table*}[t]
    \centering
    \small
    \begin{tabular}{
        p{0.13\linewidth}|
    p{0.075\linewidth}
    p{0.07\linewidth}
    p{0.07\linewidth}
    p{0.07\linewidth}
    p{0.07\linewidth}
    p{0.075\linewidth}
    p{0.07\linewidth}|
    p{0.07\linewidth}}
         & Refrigerator & Bucket & USB & Kettle & Cart & KitchenPot & Box & All\\
         \hline
        DSL-only &24.1&22.5&16.7 & 18.8 &21.4 &18.7 & 20.4& 20.6\\
        PointNet++ & 21.3 & 23.5 & 20.1& 26.6 & 24.0 & 24.4& 20.4 & 22.3\\
        MeteorNet & 33.1 & 37.4 &25.5 &35.9 & 27.0 &33.8 & 30.3 & 30.6\\
        PST-Net &20.8&21.9 &36.8 & 28.1 &31.1 &26.2& 23.5 & 27.1\\
        P4-Transformer & 29.2 &  41.2 &41.0 & 31.3 & 31.6 &34.2& 29.6 &34.5\\
        Fix+PointNet++ & 54.6 & 52.4 & 40.1 & 51.6 & 49.5 & 40.0 & 46.9 & 47.6\\
        \hline
        FixNet & \textbf{67.4} & \textbf{61.0} & \textbf{56.2} & \textbf{56.3} & \textbf{69.4} & \textbf{52.4} & \textbf{71.4} & \textbf{62.3}
    \end{tabular}%
    \vspace{-3mm}
    \caption{The accuracies of different models on \textsc{FixIt}. Our FixNet outperforms all baselines by a large margin.}
    \label{tab:acc}
    \vspace{-4mm}
\end{table*}

\noindent\textbf{Training.} The flow proposal network is pretrained on the Flyingthings3D dataset (as suggested by \cite{liu2019flownet3d}) and finetuned on 10\% ground-truth flows of videos in the training set. 
The instance segmentation network is trained on 20\% of the ground-truth segmentations of the training set using an IoU loss and a l2,1-norm regularization loss.
The physical dynamics prediction module is trained with the pseudo-labels from the flows of the videos predicted by the flow proposal network. 
The functionality prediction module is trained on the 
last frame of the simulated video of each choice, and supervised on the correctness of the five choices.

\noindent\textbf{Inference.}
Given a malfunctional object and its interaction video, we first feed the video into the flow proposal network to get the flow. Then we input the point cloud and the predicted flow into the instance segmentation network to get the instance segmentation. 

We then try to fix the object. Each choice in our choice set specifies a part index indicating the part to be modified. Retrieving from the root points, we get the root point index, and find its instance from the segmentation. We can then reconstruct the whole point cloud by selecting the points that are assigned the same instance as the root point in the choice. The DSL in each choice can be translated into a transformation matrix and applied on the part point cloud, which together with other parts constitute the fixed point cloud. With the object modified, the interacting points also need to be modified. However, the set of points to be controlled stay the same. We get a position offset as the average position change of the controlled points, and apply the same offset to our interacting points to get the revised interacting points. The revised point cloud, the segmentation  and the revised interacting points are fed into the physical dynamics prediction module to output the simulated video. We take the point cloud of the last frame and input it to the functionality prediction module to get the functionality score. The choice with the maximum score is selected.

\section{Experiments}

\subsection{Experimental Setup}

\noindent\textbf{Setup.}
The train/val/test split is approximately 6:1:3. All models select one choice out of the five candidates based on their scores. The evaluation metric is to calculate the percentage of the object instances correctly fixed by selecting the right choice.

\noindent\textbf{Baselines.}
We implement several baselines for this task. 
\setdefaultleftmargin{1em}{1em}{}{}{}{}
\begin{compactitem}
    \item \textbf{DSL-only} The DSL-only baseline takes only the choices written in domain-specific language (DSL) indicating the object after fixing is functional or not. And functionality scores are predicted based on the DSL features.

    \item \textbf{PointNet++} \cite{qi2017pointnet++} works on the single frame point cloud, which is used to examine whether the dynamics data in video assists in finding the correct fix.
    \item \textbf{MeteorNet}\cite{liu2019meteornet} adds a temporal dimension to PointNet++ to process 4D points and use chain-flow grouping.
    \item \textbf{PSTNet}\cite{fan2020pstnet} uses a point spatio-temporal (PST) convolution to represent
    point cloud sequences.
    \item \textbf{Point 4D Transformer} (P4-Transformer)\cite{fan2021point}  uses a point 4D convolution to embed the spatio-temporal local structures along with a transformer to capture the appearance and motion information.
    \item \textbf{Fix+PointNet++} Directly inputs the fixed object into PointNet++ to predict functionality.
\end{compactitem}

\noindent For the baselines taking 3D point cloud inputs (except Fix+PointNet++), we concatenate the feature of point cloud and DSL to output the functionality score of each fix.

\noindent\textbf{Implementation Details.}
For a fair comparison, all the point-cloud baselines use the segmentations from the perception module in Section \ref{sec:method}. In addition to the 3 dims representing the point positions, we add a mask dim which specifies the points to be fixed  and other points, and another dim for specifying interacting points. All the baselines use the same parameters described in the original papers, and are trained for 100 epochs. The training parameters in the individual modules of FixNet are listed in the supplementary.

\subsection{Results and Analysis}

\begin{figure*}
    \centering
    \includegraphics[width=0.9\textwidth]{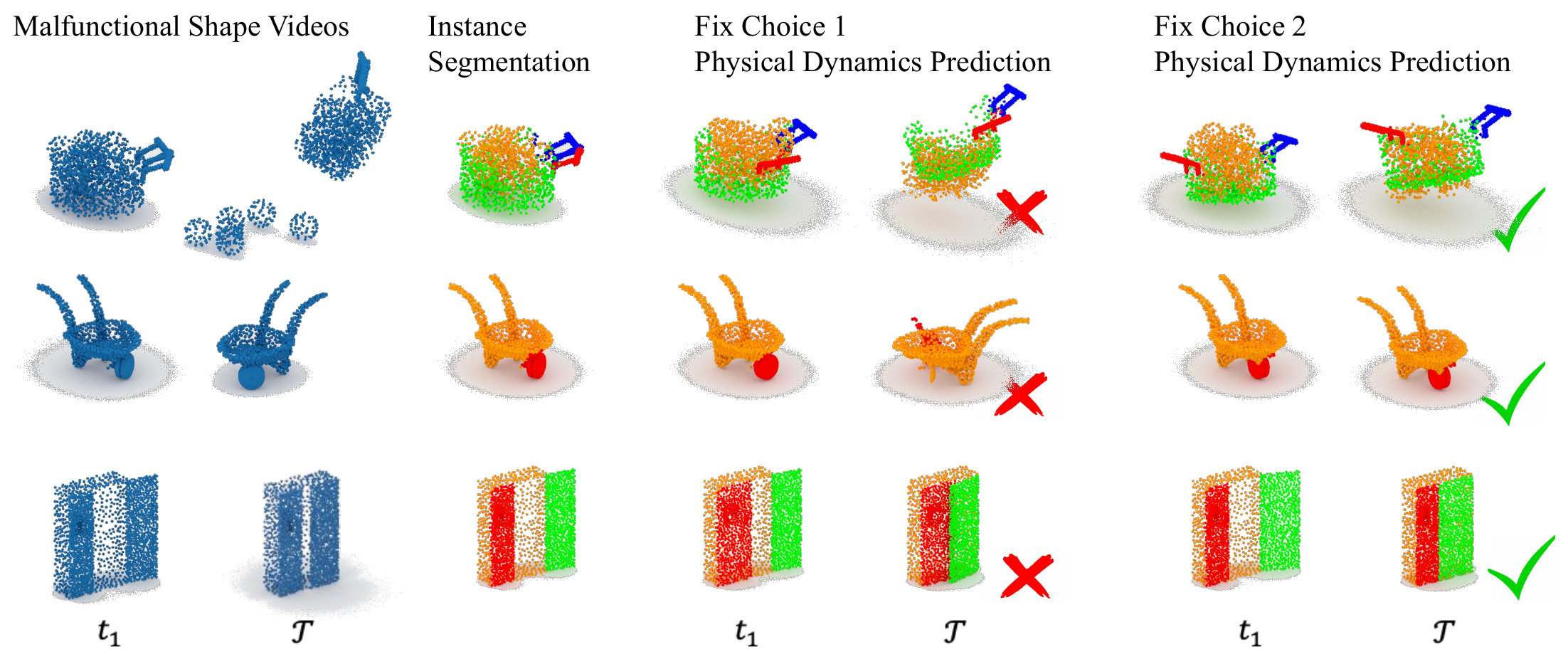}
    \vspace{-3mm}
    \caption{Qualitative examples by FixNet. Red crosses indicate unchosen fixes and green marks represent chosen fixes. As can be seen, our FixNet achieves satisfying segmentation and simulation performances.}
    \label{fig:qualitative}
    \vspace{-4mm}
\end{figure*}
\begin{table}[]
    \centering
    \resizebox{\linewidth}{!}{%
    \begin{tabular}{c|ccccccc}
    \hline
         & Fridge & Bucket & Kettle & USB & Cart & KitchenPot & Box \\
        EPE & 0.01 & 0.09 & 0.13 & 0.06 &0.03 & 0.11 & 0.02\\
        $\mathcal{L}_{IOU}$ (-) & 92.1 & 43.7 & 47.9 & 73.2 &53.5 &56.5 & 79.8 \\
    \hline
    \end{tabular}}
    \vspace{-3mm}
    \caption{Validation EPE of the flow proposal network, and IOU loss of the instance segmentation network.}
    \label{tab:epe}
    \vspace{-5mm}
\end{table}

\noindent\textbf{Main Results.} We show the multiple-choice accuracy in Table \ref{tab:acc}. As we can see, our model outperforms all baselines on point cloud video processing by a large margin. It excels in categories that involve multiple articulated parts, such as refrigerator and box.  
We also notice that for objects that are more complex in terms of physics and interactions (\textit{e.g.}, buckets, kettles and kitchenpots), the results are lower than objects with more uncomplicated physics in general. However, neural networks such as P4-Transformer and MeteorNet seem to achieve better results in these objects. The reason for this might be that the structures are more fixed for these types than others (\textit{e.g.}, the box can have arbitrary lids, but the bucket only has one handle), thus easier for neural models to memorize. FixNet is superior to Fix+PointNet++, demonstrating that physical dynamics is essential.

To get more insights about our model, we present some intermediate results on the validation set of the flow proposal network and the instance segmentation network, as in Table \ref{tab:epe}. We can see that objects with more complex physical interactions have larger errors for the flow proposal network, probably because they experience more drastic motion changes. Since the instance segmentation network is based on the flow proposal network, larger errors in the flow result in worse performances of the instance segmentation network. The inaccuracies of the perception module lead to the underperformances of the subsequent modules. How to extract precise structured representations for complex physical objects remains to be solved.

\noindent\textbf{Qualitative Examples.}
Figure \ref{fig:qualitative} shows per-module visualization results of FixNet. We can see that the perception module achieves satisfying results on segmentation for rigid body parts. However, for the kichenpot full of water, FixNet has trouble separating the water particles and the kichenpot. For the physical dynamics prediction module, FixNet can simulate different dynamics for various fixes and thus distinguishes functional fixed objects from malfunctional objects. Although the simulated dynamics are not perfect (\textit{e.g.}, some of the water leaks out of the pot, and the wheel of the cart rotates too much ), it does not prevent the functionality prediction module from predicting the right choice. 

\begin{figure}
    \centering
    \includegraphics[width=\linewidth]{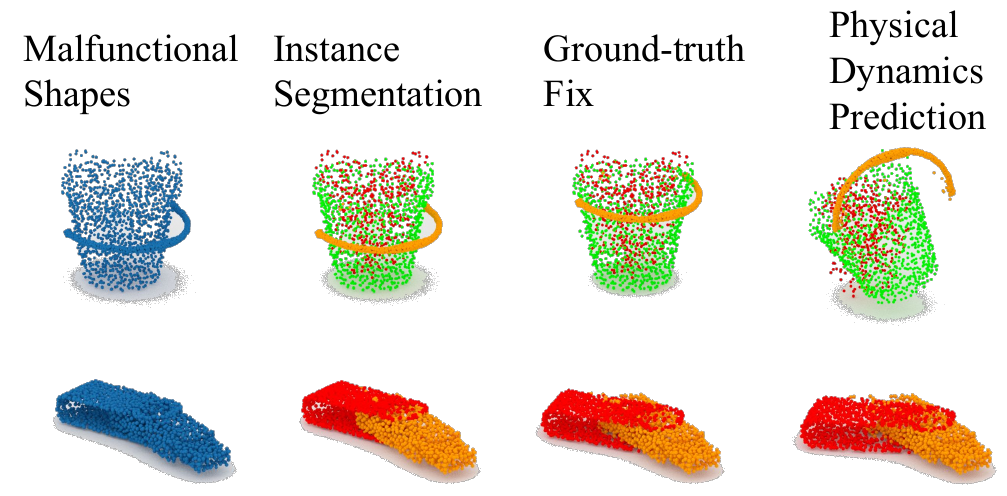}
    \vspace{-7mm}
    \caption{Failure cases by FixNet. For the bucket, the instance segmentation network cannot segment water from the bucket, and the dynamics prediction module disjoints the parts. For the USB, the overlapping of parts result in inaccurate physical simulation.}
    \label{fig:failure}
    \vspace{-8mm}
\end{figure}

\subsection{Discussions}
\noindent\textbf{Failure Cases and Challenges.}
 In Fig. \ref{fig:failure}, we show two failure cases. We discuss the limitations of our FixNet and indicate future directions for improvements below. The first weakness lies in the inaccuracy of the perception module. Specifically, DPI-Net requires the particles of different clusters to be non-overlapping. However, this is not always the case when it comes to the segmentations provided by neural networks. When two parts are close enough, overlapping inevitably occurs. Moreover, some parts are inherently embedded in others in articulated objects, making the non-overlapping requirement unrealizable. For example, the shell of the USB in Figure \ref{fig:failure} is embedded in its body, making the segmentation extremely hard. Since the hierarchical modeling in DPI-Net forces the particles in an instance to have similar dynamics, one part can be blocked by another static part due to intermediate overlapping particles. Therefore, the USB shell that is embedded in the body fails to rotate, while some USBs with separating shell and body manage to function well. In another case, we show the bucket full of water. It's hard to tell the water apart from the bucket body. We find that a large amount of the water particles are segmented into the body part. Since DPI-Net tends to unify the transformations within an instance, the bucket body might be dragged by the water when it spreads out, making the functionality prediction incorrect. As we can see, the fixed bucket should be lifted vertically, but it leans away. The second weakness is that the physical dynamics predictor fails to simulate articulated parts very well. For example, the handle of the bucket is disjointed. This also happens to the cart wheel in the qualitative examples. How to adjust dynamics model for articulated objects and objects with multiple parts is worth delving into.

\begin{figure}
    \centering
    \includegraphics[width=\linewidth]{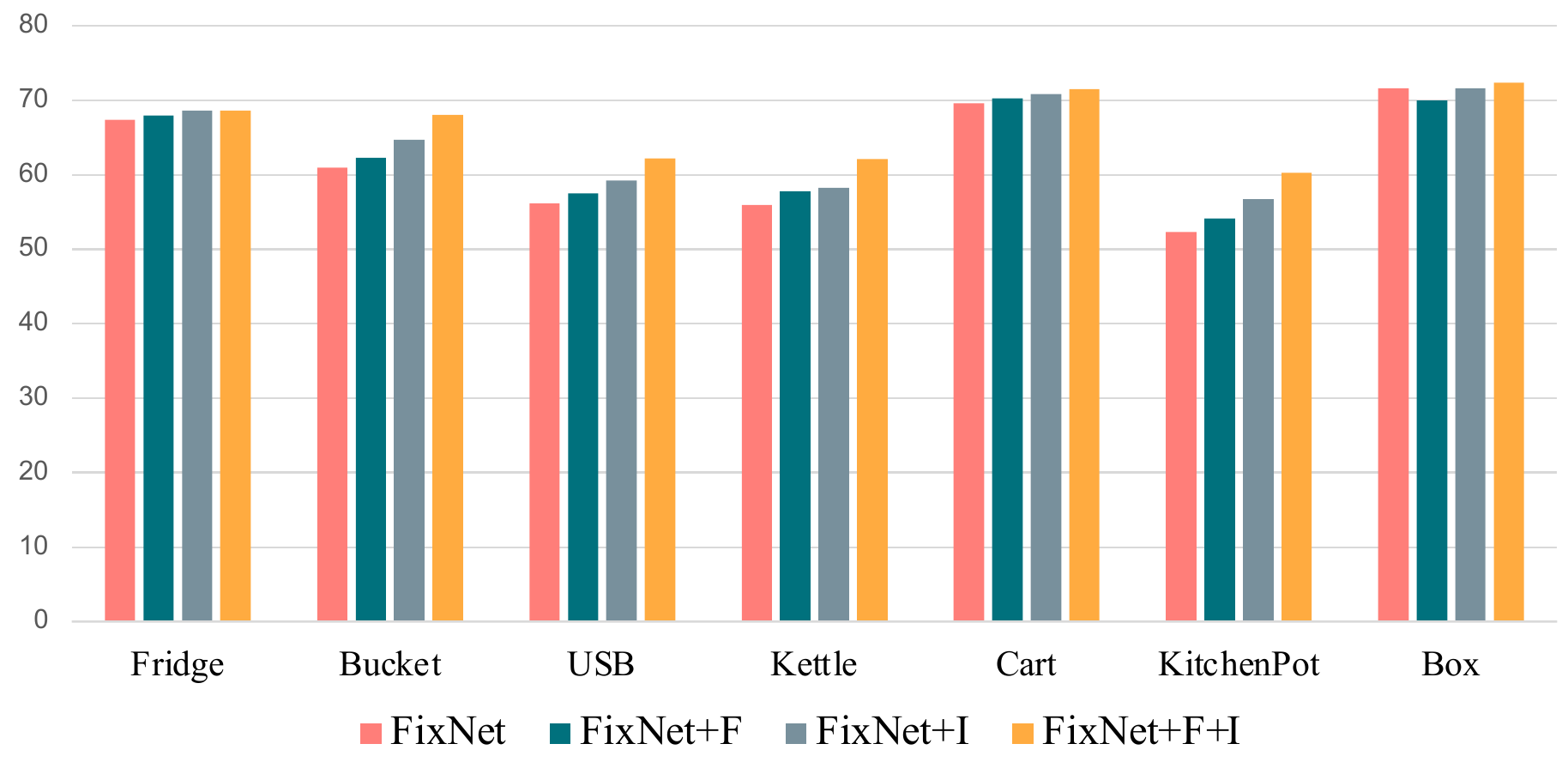}
        \vspace{-4mm}

    \caption{Model Diagnosis. Y-axis represents test accuracy. Adding ground-truth perception can boost the accuracy of FixNet to some extent, especially for categories with low accuracies.}
    \label{fig:ablative}
    \vspace{-5mm}
\end{figure}

\noindent\textbf{Model Diagnosis.}
Benefiting from our modular design, we can easily diagnose the model by replacing individual components with the ground truth data from the simulation. In Figure \ref{fig:ablative}, we show the results where we use the ground-truth flows (+F) instead of flows predicted by the flow proposal network, or ground-truth instances (+I) instead of the segmentations provided by the instance segmentation network. We can see that for categories with low accuracies by FixNet, adding the ground-truth flows or segmentations significantly improve the results. However, for categories with high accuracies, adding additional ground truths does not lead to much improvement. This suggests that the underperformances of some categories are probably due to the perception module. We notice that adding ground-truth instances leads to better performances than adding ground-truth flows. Therefore, the major challenge of this dataset might be to predict dynamics based on inaccurate segmentations. This provides insights for future explorations: improving the segmentation performances or designing a dynamics model that can take noisy perception inputs are crucial for the object-fixing task.

\subsection{Generalization}
\begin{figure}
    \centering
    \includegraphics[width=0.96\linewidth]{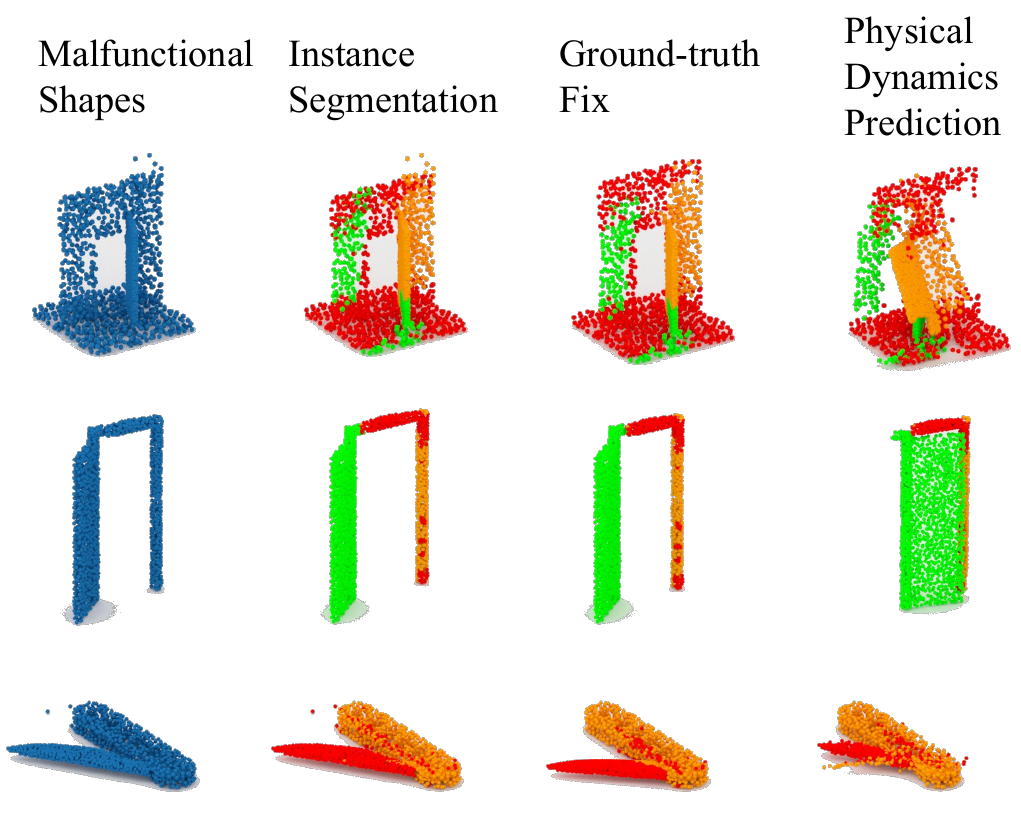}   
    \vspace{-2mm}
    \caption{Examples of generalization. For the first door, both segmentation and simulation are incorrect. For the second the door, the segmentation is incorrect but the simulation is right. For the USB, the segmentation is wrong and the simulation is half-right.}
    \label{fig:generalization}
    \vspace{-8mm}
\end{figure}

In order to evaluate our model's ability to generalize to novel categories, we conduct experiments on three unseen categories using models trained on categories with similar functionalities. Table \ref{tab:generalization} shows the generalization results. Figure \ref{fig:generalization} shows some examples of the unseen objects.

Overall, our model achieves satisfying results, outperforming P4-Transformer by a large margin. This might be credited to the generalization ability of particle-based dynamics model.  Like the mental simulation of humans, the physical dynamics predictor does not simply memorize patterns, but takes into account the physical interactions among particles. Therefore, when presented with a novel object, it is able to imagine its physical states regardless of what the object looks like. The perception module, however, does not exhibit the same generalization ability. In Figure \ref{fig:generalization}, the segmentations are  incorrect for all the three objects. For the first door, the incorrect perception poses great negative impacts on the physics module. For the second door, although the segmentation is also incorrect, the physical dynamics predictor manages to simulate the perfect results. For the USB, the model simulates the first half of the trajectory accurately, but then stops. This might be due to the slightly different interaction ways of USBs and knives. 
\begin{table}[]
    \centering
    \small
    \begin{tabular}{c|ccc}
    \hline
         & Door & Kettle (Revolute Handle) & Knife\\
         \toprule
        Train Category& Fridge & Bucket & USB\\
        \midrule
        P4-Transformer &21.3 & 24.5 & 23.0\\
        FixNet &60.4 &48.9 & 36.5\\
    \hline
    \end{tabular}
    \vspace{-2mm}
    \caption{Generalization Results. Training categories denote the categories that the models are trained on. Our FixNet shows satisfying accuracies. }
    \label{tab:generalization}
    \vspace{-4mm}
\end{table}

\section{Conclusions}

We study a novel problem of learning to fix malfunctional 3D objects 
and create a large-scale dataset \textsc{FixIt} to benchmark seven types of object functionality.
We design a novel framework FixNet that incorporates perception and physical dynamics to tackle the task.
Experiments show that our method outperforms several baseline methods.

\noindent\textbf{Limitations and Future Works.} 
We observe some failure cases when the articulated part is not well-segmented or the dynamic simulation for the articulated part and joint is inaccurate. Future works shall propose better part segmentation and dynamic models. 

\vspace{2mm}
\noindent\textbf{Acknowledgement.} This work was supported by MIT-IBM Watson AI Lab and its member company Nexplore, ONR MURI (N00014-13-1-0333), DARPA Machine Common Sense program, ONR (N00014-18-1-2847), NSF Grant BCS 1921501,  Vannevar Bush faculty fellowship, ARL grant W911NF2120104, and gifts
from MERL, Autodesk and Snap corporations.

{\small
\bibliographystyle{ieee_fullname}
\bibliography{egbib}
}

\onecolumn
\appendix

\begin{center}
	{
		\Large{\textbf{Supplementary Materials for   ``Fixing Malfunctional Objects With  Learned Physical Simulation and Functional Prediction''}}
	}
\end{center}

\vspace{15 pt}

\begin{itemize}[leftmargin=*]
    \item In Section~\ref{sec:dataset}, we provide more details on the dataset.
    \item In Section~\ref{sec:implementation}, we provide more details on the experimental setup
    
    \item In Section~\ref{sec:result}, we show more experimental results.
\end{itemize}

\section{Dataset}
\label{sec:dataset}
\begin{figure}[htbp]
    \centering
    \includegraphics[width=\linewidth]{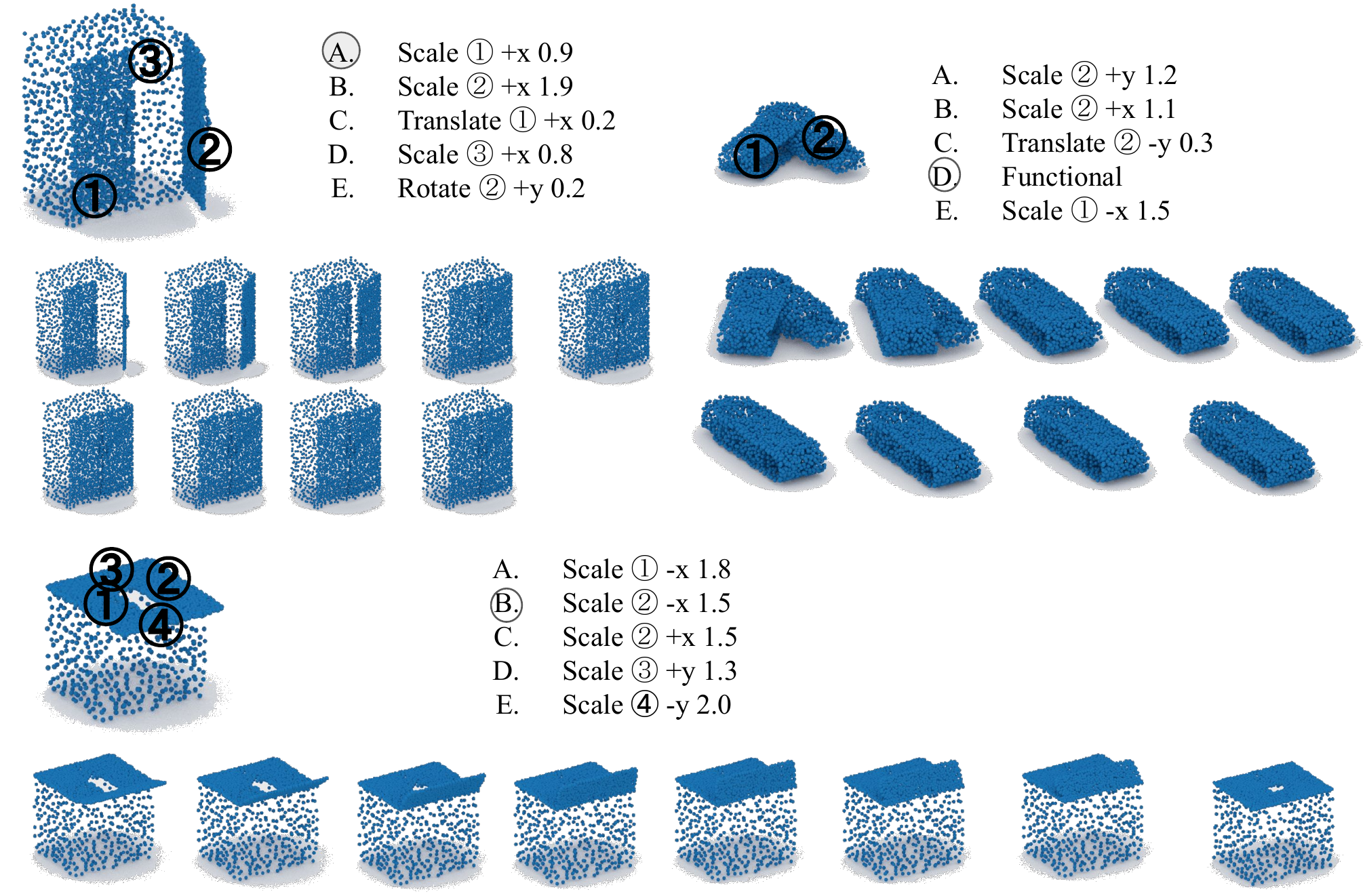}
    \caption{Examples of our \textsc{FixIt} dataset.}
    \label{fig:exp}
\end{figure}

In Table \ref{tab:type}, we show the fixing types (scale, translate, rotate) that can be applied to each category and included in the choices.
\begin{table}[htbp]
    \centering
    \begin{tabular}{c|ccccccc}
    \hline
        Category & Refrigerator& Bucket & USB & Kettle & Cart & KitchenPot & Box \\
        Fixing Type & S, T, R & S, T & S, T & R & R & R & S\\
    \hline
    \end{tabular}
    \caption{The fixing types that can be applied to the categories. S=Scale, T=Translate, R=Rotate}
    \label{tab:type}
\end{table}
In Figure \ref{fig:exp}, we add more examples of our dataset.
\section{Implementation Details}
\label{sec:implementation}
\subsection{DSL-based baselines}

For the DSL-based baselines, we input the ``type" and ``axis" into an embedding layer and then pass through a LSTM layer to extract language features. The value is fed into a linear layer and concatenated with the language features. The DSL features go through MLP to output binary classification results.

\subsection{FixNet}
For the flow proposal network, we pretrain FlowNet3D on the Flyingthings3D\cite{mayer2016large} dataset, and finetuned on 10\% of the ground-truth flows of the videos in our training set. We use a learning rate of 0.001, batch size of 32, and dropout of 0.5. We train the model for 250 epochs, and use the model that achieves best performance on the validation dataset. 

For the instance segmentation network, we use a learning rate of 0.001, decay rate of 1e-4 for weights, decay steps for learning rate as 20 and decay rate of 0.5 for learning rate. The batch size is 16. We use Adam Optimizer to train for 250 epochs. 

The physical dynamics prediction module is trained for 1.5M iterations till converge using Adam optimizer with initial learning rate 1e-4.

For the funcitonality prediction network, we use learning rate of 0.001 and decay rate of 1e-4 to train for 200 epochs. The batch size is 24. 

\section{More Qualitative Examples}
\label{sec:result}
\begin{figure}
    \centering
    \includegraphics[width=\textwidth]{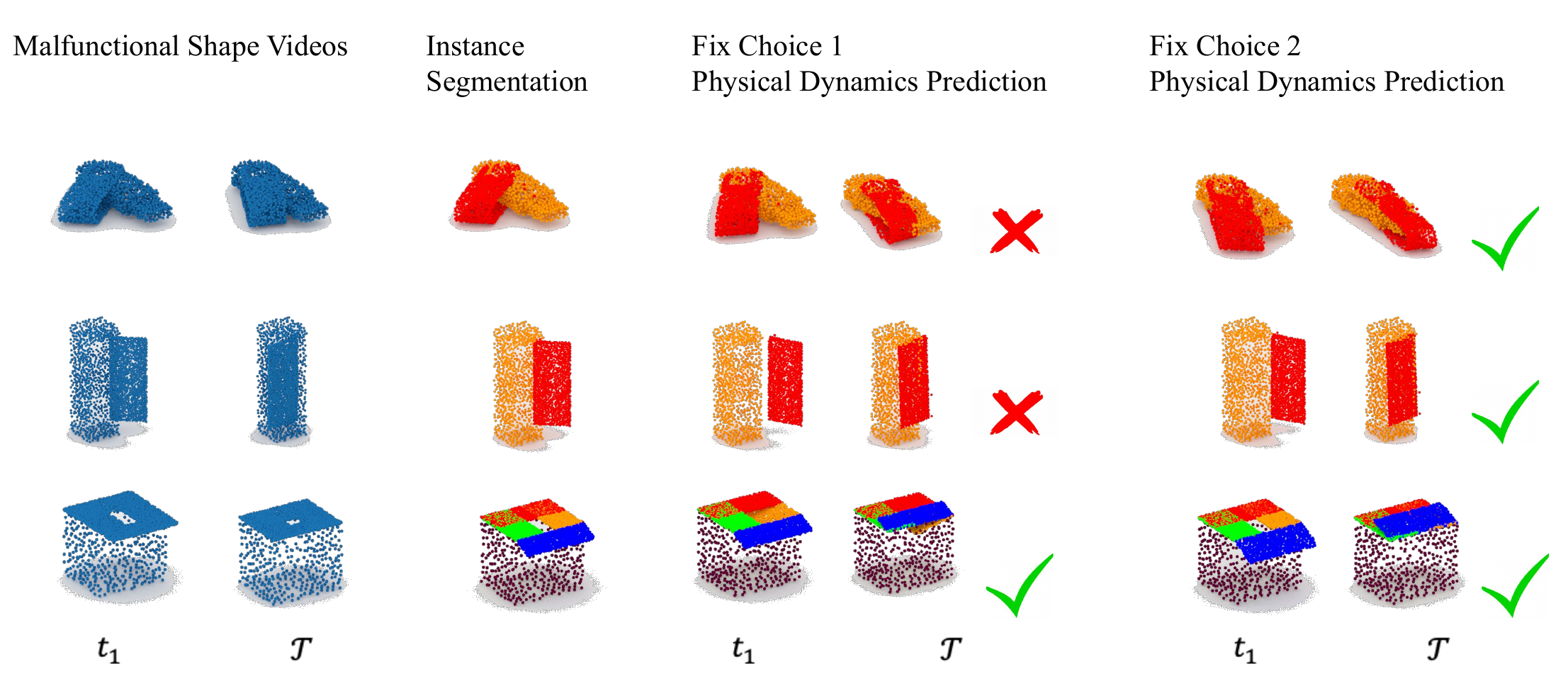}
    \caption{Qualitative Examples of FixNet}
    \label{fig:qualitative2}
\end{figure}

In Figure \ref{fig:qualitative2}, we show some more qualitative examples output by FixNet.

\end{document}